\pdfoutput=1
\documentclass[11pt]{article}

\usepackage[final]{acl}

\title{FOL2NS: Generating Natural Sentences from First-Order Logic}
\author{Mei Jia $^{\dagger}$\thanks{~~This work was completed during the author's Master's studies at the University of Manchester.}\\
  University of Manchester\\
  \texttt{jiamei.meow@gmail.com} \\}

\usepackage{times}
\usepackage{latexsym}
\usepackage{array}
\usepackage{booktabs}
\usepackage{multirow}
\usepackage[T1]{fontenc}
\usepackage[utf8]{inputenc}
\usepackage{microtype}
\usepackage{inconsolata}
\usepackage{graphicx}
\usepackage{textgreek} 
\usepackage{float}
\usepackage[section]{placeins}
\usepackage{amsmath}
\usepackage{amssymb}
\usepackage{multicol}
\usepackage{xurl}    

\begin{document}
\maketitle
\begin{abstract}
Translating formal language into natural language is a foundational challenge in NLP, driving various downstream applications in semantic parsing, theorem validation, and question-answering. In this study, we introduce First-Order Logic to Natural Sentence (FOL2NS), a neurosymbolic framework designed to generate synthetic FOL formulas and convert them into natural human expressions. It handles deeply nested structures with varying quantifier depths (QD), which are rarely captured by existing corpora. By combining rule-driven modules with fine-tuned language models, FOL2NS enhances the diversity and coverage of the generated samples. In our experiments, we systematically evaluate the framework's capabilities through both character-level analysis and overall performance metrics. Experimental results show that FOL2NS can reliably produce well-formed templates and fluent statements, but it faces challenges in achieving precise semantic representations and natural generation as structural complexity increases.
\end{abstract}
  
\section{Introduction}
In recent years, Large Language Models (LLMs) have shown impressive performance in Natural Language Generation (NLG). Despite this progress, their ability to tackle logic-to-text tasks requires more exploration, especially when generating natural sentences from first-order logic (FOL). The demand for formal correctness is equally important as fluent, coherent texts in such tasks. To this end, we propose a novel FOL2NS translation pipeline based on formal syntax, which bridges the rigour of the symbolic system and the data-driven power of large pre-trained models. A related example of outputs in every step is shown in Table~\ref{tab:eg_FOL2NS}.

The pipeline construction comprises multiple steps: (i) generating mathematical formulas by context-free grammars, (ii) incorporating real-world lexical items into the formulas to get their logical forms, (iii) finetuning a pretrained T5 model on a preprocessed dataset from an existing corpus, and (iv) mapping the logic forms to natural sentences via this model. In theory, this structure can generate unlimited formulas and increase the overall sample size under explicit changes of length and depth. It enables on-demand adjustment of the maximum nesting level and seamless integration of new templates into the pipeline. Simultaneously, it facilitates precise control over the complexity of both generated FOL formulas and their corresponding natural sentences. Related source code is available at:\url{https://github.com/Mifuxuanan/LELA70502_FOL2NS-Generating-Natural-Language-Sentences-from-First-Order-Logic}.

\section{Related Work}
Previous work has connected natural language with logical representations and provides parallel or semantically aligned translation forms. On one hand, for NLG from structured tables, \citet{chen-2020-logic2text} introduces Logic2Text, a large-scale dataset of 10,753 paired logical forms and descriptions. They employ several models from rule-based templates, encoder-decoder architectures, to fine-tuned GPT-2, for controllable and high-fidelity NLG. Differently, my work focuses on generating texts from first-order logic (FOL) through fine-tuning the T5 model with different strategies.

\begin{table*}[htbp]
  \centering
  \small
  \renewcommand{\arraystretch}{1.2}
  \begin{tabular}{
    p{3cm}  p{12cm}}
    \toprule
    \textbf{Type} & \textbf{Output}\\
    \midrule
        \addlinespace[0.5ex]
    FOL Formula  & $\forall a\bigl(A(a)\to\forall b(\neg B(b,c)\lor\neg C(b))\bigr)$ \\
    Lexicalized FOL  & $\forall a(HasOfficeIn(a)\to\forall chef(\neg LivesIn(chef, zone)\lor\neg IsThoughtful(chef)))$\\
    Natural Sentence  & If the chef has an office in a building, then the chef neither lives in the zone nor is thoughtful. \\
    \bottomrule
  \end{tabular}
  \caption{An output example in every step of constructing FOL2NS. It starts with an FOL formula with the quantifier depth of two in both mathematical and lexicalised formats. The corresponding natural sentence is generated by fine-tuning the T5-large model.}
  \label{tab:eg_FOL2NS}
  \end{table*}

On the other hand, there are several studies about Natural Language(NL) to FOL translation. LogicNLI \citep{tian2021} offers a large-scale natural language inference (NLI) dataset for multi-step FOL reasoning with 20k semi-synthetic examples. FOLIO \citep{han2024folionaturallanguagereasoning} contains 1,430 human-annotated natural language reasoning examples with verified FOL translations. NL2FOL \citep{lalwani2025} presents a neurosymbolic pipeline that autoformalizes natural language arguments into FOL through LLM-driven decomposition and SMT validation. However, these studies only exhibit the logical form as the translation outputs and don't consider the nesting levels controlled by quantifiers in FOL formulas. In contrast, my work starts with the initial mathematical formulas and highlights every step of how to map rules with quantifier depths to natural sentences. It also covers the T5 model’s text generation quality from multiple nested logical structures.

\section{Definition}\label{chap:3}
\subsection{First-order logic} \label{sec:3.1}
First-order logic (FOL) is a logical system for reasoning about the properties of objects or the relationships between objects in a domain. This kind of proposition consists of k-ary predicate symbols, variables, constants and function symbols. In FOL, the atomic formulas (e.g. $P(a)$, $Q(a,b)$) are predicates that assert a relationship among certain elements. Beyond that, compound formulas are assembled using more logical connectives, such as negation ($\neg$), implication ($\to$), exclusive-or ($\oplus$), conjunction ($\land$) and disjunction($\lor$). 

\subsection{Quantifier Depth} \label{sec:3.2}
The most significant concept in FOL is quantification: the ability to assert that a certain property holds for all elements with universal quantifiers ($\forall$) or that it holds for some element with existential quantifiers ($\exists$). "Quantifier" is the key feature that sets FOL and other propositional logics apart. To measure the structural complexity of an FOL formula, one often considers its quantifier depth (QD). Given the formula F and formula G, the related relationships are defined as follows:\\
    \hspace*{1.5em}1. An atomic formula has $QD = 0$;\\
    \hspace*{1.5em}2. Negation does not increase the quantifier depth: $QD(\neg F) = QD(F)$;\\
    \hspace*{1.5em}3. For binary connectives, the quantifier depth is the maximum depth between the sides of operands: $QD(F\land G) = QD(F\lor G) = max(QD(F), QD(G))$;\\
    \hspace*{1.5em}4. Each application of a quantifier increases the depth by one: $QD(\forall xF) = QD(\exists xF) = QD(F)+1$.

\section{Data Construction}
\begin{figure*}[htbp]
\includegraphics[width=1\linewidth]{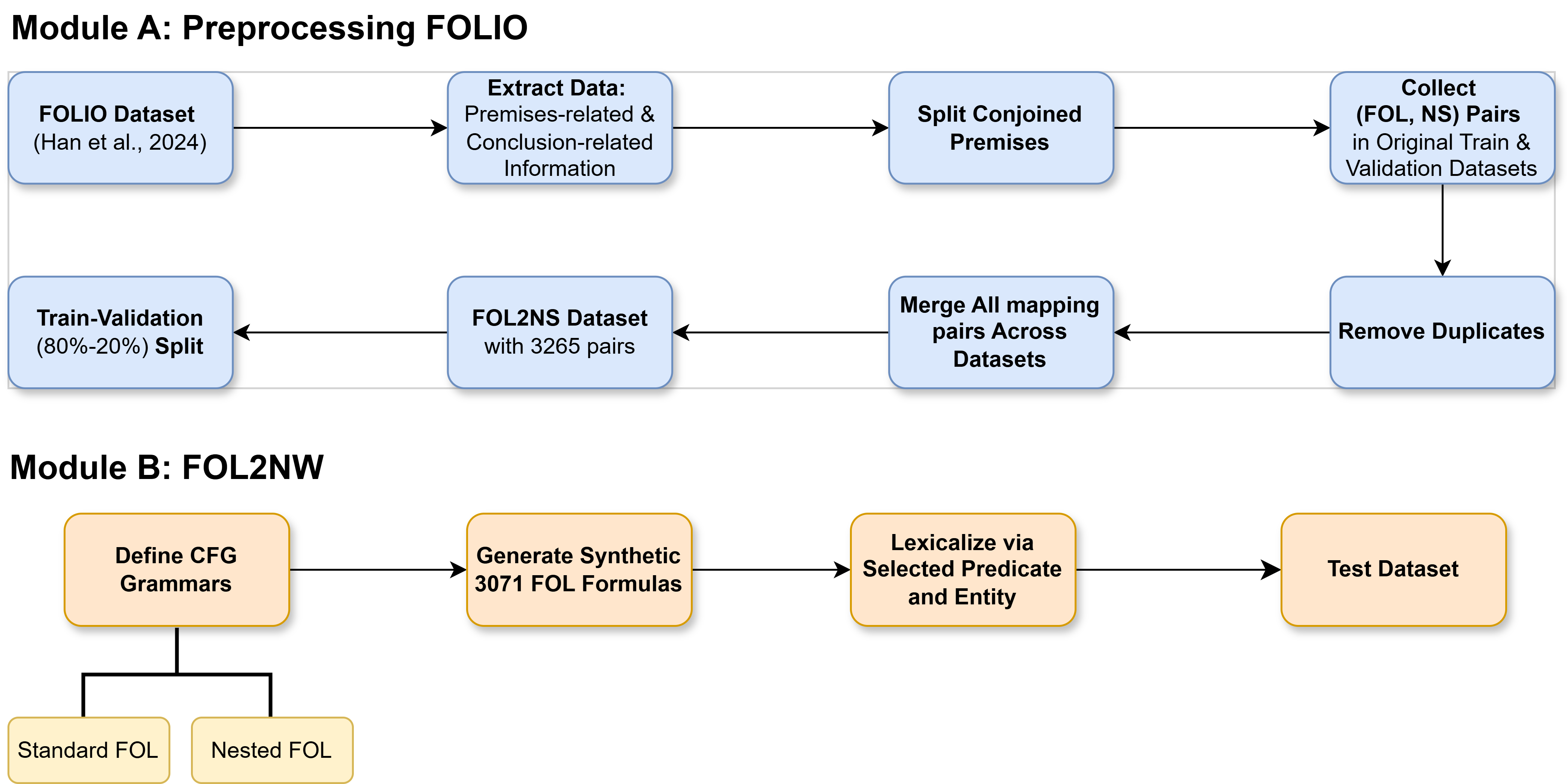} 
\hfill
\caption {Overview of the proposed framework for dataset construction. Module A gets the Train and Validation datasets by cleaning the FOLIO dataset. Module B generates the Test dataset by CFG rules and FOL-to-Natural-Words mapping.}
\label{tag:pipeline}
\end{figure*}

The dataset for fine-tuning models is constructed in 2 stages as illustrated in Figure~\ref{tag:pipeline}. The source of Training and Validation sets is FOLIO \citep{han2024folionaturallanguagereasoning}, an expert-written dataset designed for FOL reasoning. Despite its rich logical and semantic complexity, it has a distinct limitation in the sample size. Based on the original 1,430 samples of FOLIO, we expanded thousands of test samples through synthetic FOL formulas. This approach combines the strengths of both to systematically construct a diverse and controllable dataset.
\subsection{FOLIO Dataset} 
\citet{han2024folionaturallanguagereasoning} present a novel NL-to-FOL translation benchmark: FOLIO. It is inherently equipped with parallel pairs between natural-language statements and their FOL representations. Every premise and conclusion pair has been transcribed into FOL format and verified by an FOL inference engine, ensuring logical correctness. Compared to other synthetic datasets, each example in FOLIO is mostly aligned with a real-world background, with increased reasoning complexity and a larger vocabulary size of 4351. 

FOLIO dataset contains 1,001 training and 203 validation instances drawn from 487 original “stories” with seven columns, including paired logical and NL representations of premises and conclusions, three classification labels and information about their ID (Appendix Figure~\ref{tag:FOLIO_eg}). It also involves various logic patterns with special symbols ($\forall$, $\neg$, $\land$, $\lor$). However, the cases of existential quantification ($\exists$) and initial FOL formulas are not present in this dataset, which are complemented in the Test dataset of FOL2NS. 

\subsection{Module A: Preprocessing FOLIO} 
\begin{figure}[htbp]
\centering
  \includegraphics[width=0.8\linewidth]{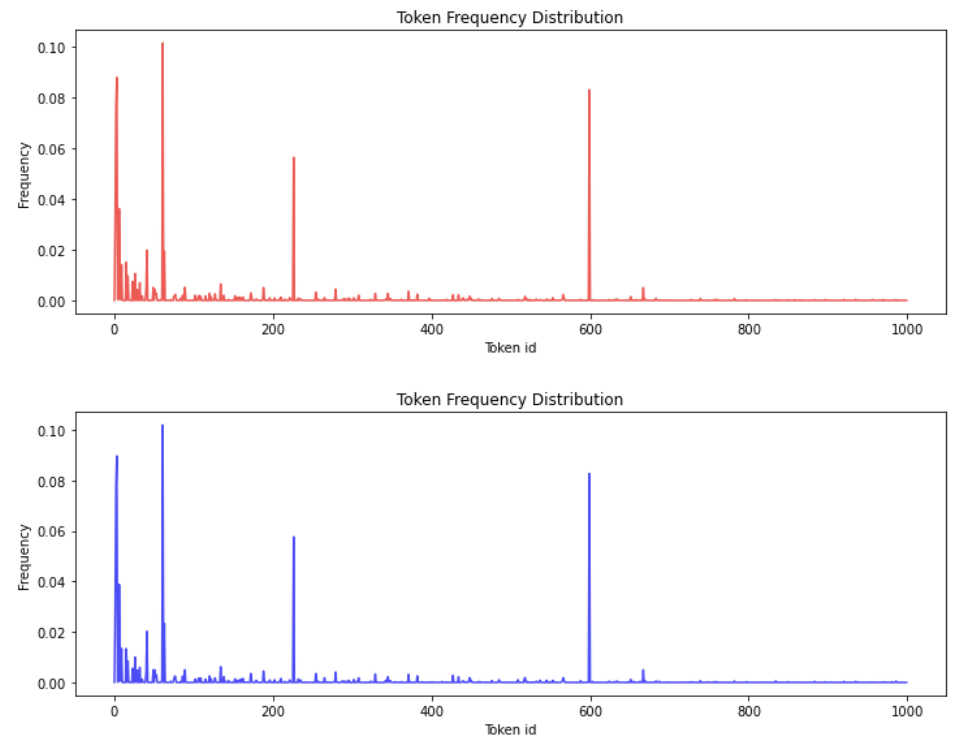} 
  \hfill
  \includegraphics[width=0.8\linewidth]{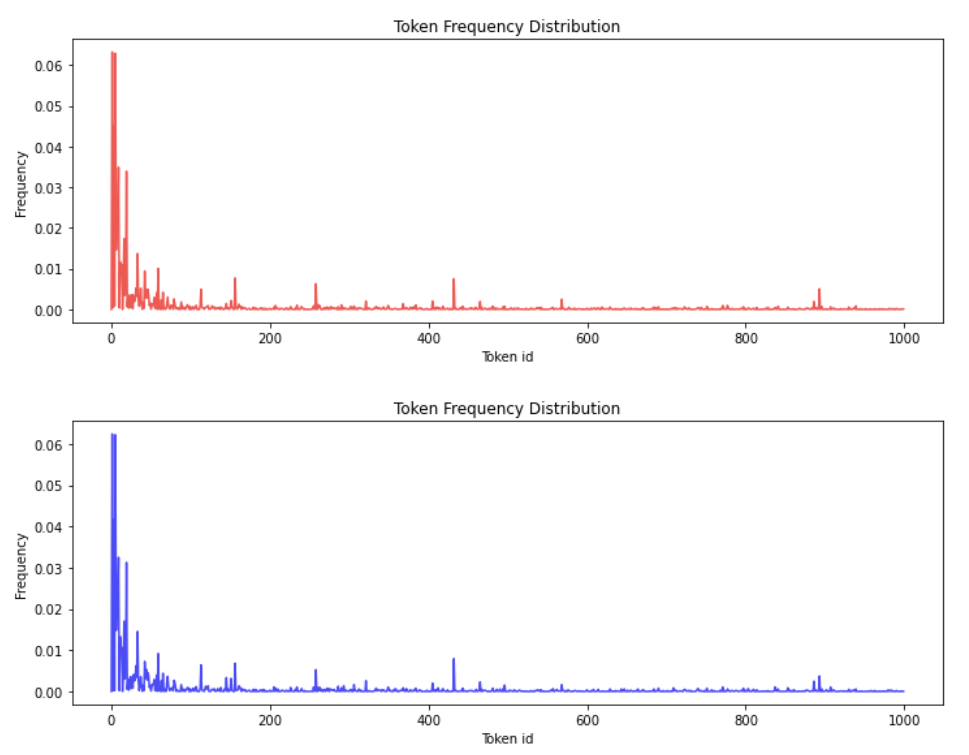}
  \caption {The token frequency distribution of FOL as inputs (above two) and NS as outputs (bottom two), for the preprocessed dataset of Module A in Training (red) and Validation (blue) splits.}
\label{tag:newFOLIO_dist}
\end{figure}

The data cleaning of FOLIO is conducted through several steps. We began with extracting the premise-related and conclusion-related components from each sample. The conjoined premises were split into individual clauses so that each natural sentence (NS) aligned clearly with a single logical form. Then, by reversing the mapping order in FOLIO, we collected all resulting pairs of FOL and NS from the original datasets. After removing duplicate pairs, this process yielded 3,625 formula-statement pairs and was finally partitioned into an 80\%-20\% Training-Validation split for subsequent fine-tuning. The related statistical information is shown in Figure~\ref{tag:newFOLIO_dist} and Table~\ref{tab:T5_KL}. The token frequency distributions for inputs (FOL) share an analogous shape as those for outputs (NS). In all cases, the smaller Kullback-Leibler divergence $D_{KL}(validation||train)$ ensures that the model can see both high-frequency tokens and tail-end vocabulary across splits for higher-quality generation.

  \begin{table*}[htbp]
  \centering
  \small
   \renewcommand{\arraystretch}{1.2}
  \begin{tabular}{ 
    >{\centering\arraybackslash}p{4.5cm} >{\centering\arraybackslash}p{2cm} >{\centering\arraybackslash}p{2cm}
  }
    \toprule
    \multirow{2}{*}[-1ex]{\bfseries Train (P) vs. Validation (Q)}
      & \multicolumn{2}{c}{\bfseries KL Divergence}\\
    \cmidrule(lr){2-3}
      &{\bfseries PQ}
      & {\bfseries QP} \\
    \midrule
    \addlinespace[0.5ex]
    Input  & 0.5094 & 0.2186\\
    Output  & 1.1563 & 0.4896\\
    \addlinespace[0.5ex]
\bottomrule
  \end{tabular}
  \caption{KL divergence of Train versus Validation sets in Module A.}
 \label{tab:T5_KL}
  \end{table*}

 \begin{table*}[htbp]
  \centering
  \small
  \renewcommand{\arraystretch}{1.2}
  \begin{tabular}{
    p{5cm}  p{5cm}  }
    \toprule
    \textbf{Symbol} & \textbf{Lexical Item}\\
    \midrule
        \addlinespace[0.5ex]
    Original Formula & $\forall b(\neg B(b, c))$ \\
    Sampling Predicate and Related Entity & $LivesIn([$"$Person$", "$Location$"$])$ \\
    Sampling Term in the Target Entity & "$Person$"$:\ chef$, "$Location$"$:\ zone$\\
    Final Output & $\forall chef (\neg LivesIn(chef, zone))$ \\
    \bottomrule
  \end{tabular}
  \caption{A simplified example of FOL2NW workflow.}
  \label{tab:process}
  \end{table*}
  
\subsection{Module B: FOL2NW} 
Constructing the test data involves creating a synthetic set of FOL expressions and converting them into the lexicalised logical form. To increase structural complexity, the generation strategy contains two context-free grammars (CFGs) over the logical connectives and atomic predicates. The design idea is based on the definition of FOL (§~\ref{sec:3.1}) and QD (§~\ref{sec:3.2}). For “Standard CFG”, it started from atomic formulas and repeatedly combined logical operators from left to right, resulting in flat, two-way branches. By adjusting the maximum depth, it was generated with increasing length but without nested predicate arguments. In addition to the above, “Nested CFG” permitted predicate symbols to take other formulas as arguments. Again, a global depth parameter limits term-argument nesting, which allows the control of quantifier depth and formula complexity at the same time. To align with the logic formatting used in FOLIO, we constrained the whole depth in the range of 4-10. We obtained 3,071 unique FOL formulas by uniformly sampling from both grammars. The related examples are shown in Appendix Figure~\ref{tag:eg_FOL}. 

The next step is to fill each abstract formula with lexical items and convert them into concrete, readable formats. First, we defined two vocabularies: a set of 50 predicate names, including both unary predicates (e.g., “$IsHappy$”) and binary predicates (e.g., “$Like$”), and a set of entities under seven classes (“$Person$”, “$Organization$”, “$Location$”, “$Field$”, “$Object$”, “$Animal$”, and “$Drink$”). Second, the lexicalised procedure was performed for every CFG-based formula. We uniformly sampled from the 50 predicates, and according to their arities, selected the random individual term in the related entity class. Third, each variable in the template was substituted by the corresponding sampled predicate or entity, except first variables did not change to keep the subject-predicate agreement in the statement. A simplified step-by-step example is displayed in Table~\ref{tab:process}. By repeating the whole operation for all 3,071 samples, it produces a diverse test set of fully lexicalised FOL logical format. The related examples are shown in Appendix Figure~\ref{tag:T5_result}.

\section{Experimental Setups}
This section includes additional task-relevant symbols to the vocabulary of pretrained T5, the details of fine-tuning setup and metrics for evaluating model performance with character-level Levenshtein Distance and adjusted BLUE score.
\subsection{Model}
We selected vanilla pretrained T5 \citep{raffel2023exploringlimitstransferlearning} to fine-tune on the prepared dataset in Module A and tested it on the generated logical formats in Module B. Due to its text-to-text paradigm, the model allows us to take FOL as the source language and English as the target language without any additional architectural change. During inference, the flexible decoding strategies, such as beam search and repetition penalty, provide my work with various methods for high-quality generation. 

Although T5 has been pre-trained on massive amounts of data, its original vocabulary does not contain tokens of logical operators and quantifiers. We tried to directly add all these special symbols in the tokenising stage, but the model failed to recognise them and still needs enough training on this task. Therefore, we decided to replace these symbols with the corresponding lexical items referring to Table~\ref{tab:symbol}.

\begin{table}[htbp]
  \centering
  \small
  \renewcommand{\arraystretch}{1.2}
  \begin{tabular}{
    >{\centering\arraybackslash}p{2cm}    
    >{\centering\arraybackslash}p{2cm}  }
    \toprule
    \textbf{Symbol} & \textbf{Lexical Item}\\
    \midrule
        \addlinespace[0.5ex]
    $\neg$ & No \\
    $\forall$ & For All \\
    $\exists$ & There Exists \\
    $\oplus$ & XOR \\
    $\to$ & implies \\
    $\land$ & and \\
    $\lor$ & or \\
    \bottomrule
  \end{tabular}
  \caption{The mapping relationship between symbols and lexical items.}
  \label{tab:symbol}
  \end{table}

\subsection{Fine-tuning Details}
We consider different optimisation-related hyperparameters, including learning rates at 1e-4 and 1e-5, and training epochs of 20, 25 and 30. To ensure high-quality generation without excessive repetition or unnecessary parts, we also applied four decoding strategies: “Standard”, “Adjusted”, “Prefixed” and “Adjusted with Length”. Each strategy represented a further optimisation built upon the previous one. For “Standard”, we removed every padding token’s label by $-100$ to avoid appearing “$<pad>$” in outputs. For “Adjusted”, we changed the arguments in “$model.generate$” with “$early\_stopping$”, “$num\_beams$” as 5, “$repetition\_penalty$” as 1, and “$no\_repeat\_ngram\_size$” as 2. For “Prefixed”, we added a prefix “Translate FOL formula to English:” to clarify the objective of the model. For “Adjusted with Length”, we set “max length” from the default value of 20 to 64, which prevented the model from being prematurely truncated. Given training efficiency and cost, we tried T5-base with a batch size of 32 in initial experiments and compared different configurations. However, we finally decided to go ahead with T5-large with a batch size of 8 to gain better performance. 

\subsection{Performance Metrics}
We used Levenshtein Distance and BLEU score for overall evaluation in the Validation stage. Shorter distance indicates closer alignment to the reference, while BLEU scores range from 0 to 1, with values closer to 1 representing higher generation quality. The related notations are defined as follows: for each valid example $i$, the model’s generated sentence is viewed as candidate $\hat{y_i}$, and its reference label is $y_i$. The number of all non-empty pairs is denoted as $N$. \\

\textbf{Levenshtein Distance.} \citet{levenshtein1966} defined it as a string metric for measuring the difference between two sequences. Building on the standard definition, we apply a character-level edit distance metric, and denote $D(\hat{y_i}, y_i)$ as the minimum edit number to convert $\hat{y_i}$ into $y_i$. Beyond raw edit counts, the distance score is normalised by the maximum token length of the pair $(\hat{y_i}, y_i)$. The Levenshtein distance $d_i$ and related score $s_i$ are calculated as follows:\\
\hspace{-10pt}
\begin{equation}
  d_i = D(\hat y_i, y_i)
  \label{eq:di}
\end{equation}

\begin{equation}
  s_i = \frac{D(\hat y_i, y_i)}
             {\max\bigl(|\hat y_i|,\;|y_i|\bigr)}
  \label{eq:si}
\end{equation}

The average distance $\bar{D}$ and average normalized score $\bar{S}$ are reported as follows:\\
\begin{equation}
  \bar{D}
  = \frac{1}{N}\sum_{i=1}^N d_i
  = \frac{1}{N}\sum_{i=1}^N D(\hat y_i,y_i)
  \label{eq:dbar}
\end{equation}

\begin{equation}
  \bar{S}
  = \frac{1}{N}\sum_{i=1}^N s_i
  = \frac{1}{N}\sum_{i=1}^N
    \frac{D(\hat y_i,y_i)}
         {\max\bigl(|\hat y_i|,\;|y_i|\bigr)}
  \label{eq:sbar}
\end{equation}

\textbf{BLEU Score.} We adopt the original BLEU formulation of \citet{BLEU2002}, but introduce a few adjustments to make it more robust in practice. The outer summation over multiple references reduces to a single sentence, as my case has only one gold label $y_i$ per candidate $\hat{y_i}$. To keep numerical stability, we also add an epsilon $\epsilon$ of $10^{-3}$ to the denominator in clipped-count normalisation and brevity penalty. The related formulas are calculated as follows:\\
\begin{equation}
  p_n
  = 
  \frac{\displaystyle
        \sum_{n\text{-gram}\,\in\,\mathrm{cand}}
           \mathrm{Count}_{\mathrm{clip}}(n\text{-gram})}
       {\displaystyle
        \sum_{n\text{-gram}'\,\in\,\mathrm{cand}}
           \mathrm{Count}(n\text{-gram}') + \epsilon}
  \label{eq:pn}
\end{equation}

\begin{equation}
  \mathrm{BLEU}_i
  = \mathrm{BP}_i\,
    \exp\!\Bigl(\frac{1}{N}\sum_{n=1}^N \ln p_{n,i}\Bigr)
  \label{eq:bleui}
\end{equation}

\begin{equation}
  \mathrm{BP}
  =
  \begin{cases}
    1,                                         & c > r,\\[4pt]
    \exp\!\Bigl(1 - \tfrac{r}{c + \epsilon}\Bigr), & c \le r,
  \end{cases}
  \quad \epsilon = 10^{-3}
  \label{eq:bp}
\end{equation}

The average BLEU score $\overline{\mathrm{BLEU}}$ is reported as follows:\\
\begin{equation}
  \overline{\mathrm{BLEU}}
  = \frac{1}{N}\sum_{i=1}^N \mathrm{BLEU}_i
  \label{eq:bleu_avg}
\end{equation}

\begin{table*}[htbp]
  \centering
  \small
   \renewcommand{\arraystretch}{1.2}
  \begin{tabular}{ 
    >{\centering\arraybackslash}p{3cm} 
    >{\centering\arraybackslash}p{1.5cm} >{\centering\arraybackslash}p{1.5cm} >{\centering\arraybackslash}p{1.5cm} >{\centering\arraybackslash}p{3cm}
  }
    \toprule
    \multirow{2}{*}[-1ex]{\bfseries Metric}
      & \multicolumn{4}{c}{\bfseries Training Strategy}\\
    \cmidrule(lr){2-5}

      & {\bfseries Standard}
      & {\bfseries Adjusted}
      & {\bfseries Prefixed} 
      & {\bfseries Adjusted with Length} \\
    \midrule
    \addlinespace[0.5ex]
    Avg Edit Distance  & 31.99 & 27.83 & 26.98 & 22.97 \\
    Avg Distance Score  & 2.42 & 2.13 & 1.94 & 1.71  \\
    AVG BLEU  & 0.44 & 0.47 & 0.52 & 0.59 \\
    \addlinespace[0.5ex]
\bottomrule
  \end{tabular}
  \caption{Performance metrics with a learning rate of $1e-4$ using different training strategies.}
  \label{tab:diff_strategy}
  \end{table*}
\begin{table}[htbp]
  \centering
  \small
   \renewcommand{\arraystretch}{1.2}
  \begin{tabular}{ 
    >{\centering\arraybackslash}p{3cm} 
    >{\centering\arraybackslash}p{1.5cm} >{\centering\arraybackslash}p{1cm} >{\centering\arraybackslash}p{1cm} >{\centering\arraybackslash}p{1cm}
  }
    \toprule
    \multirow{2}{*}[-1ex]{\bfseries Metric}
      & \multicolumn{2}{c}{\bfseries Learning Rate}\\
    \cmidrule(lr){2-3}

      & {\bfseries 1e-4}
      & {\bfseries 1e-5} \\
    \hline
    \addlinespace[0.5ex]
    Avg Edit Distance  & 27.83 & 44.78\\
    Avg Distance Score  & 2.13 & 3.93 \\
    AVG BLEU  & 0.47 & 0.092\\
    \bottomrule
  \end{tabular}
  \caption{Performance metrics across different learning rates using the Standard training strategy.}
  \label{tab:diff_lr}
  \end{table}

\begin{table}[htbp]
  \centering
  \small
   \renewcommand{\arraystretch}{1.2}
  \begin{tabular}{ 
    >{\centering\arraybackslash}p{3cm} 
    >{\centering\arraybackslash}p{1cm} >{\centering\arraybackslash}p{1cm} >{\centering\arraybackslash}p{1cm}
  }
    \toprule
    \multirow{2}{*}[-1ex]{\bfseries Metric}
      & \multicolumn{3}{c}{\bfseries Training Epoch}\\
    \cmidrule(lr){2-4}
      &{\bfseries 20}
      & {\bfseries 25}
      & {\bfseries 30} \\
    \midrule
    \addlinespace[0.5ex]
    Avg Edit Distance  & 33.46 & 28.69 & 27.83 \\
    Avg Distance Score  & 2.64 & 2.21 & 2.13  \\
    AVG BLEU  & 0.36 & 0.45 & 0.47  \\
    \addlinespace[0.5ex]
\bottomrule
  \end{tabular}
  \caption{Performance metrics across different training epochs with a learning rate of $1e-4$ using the Adjusted training strategy.}
  \label{tab:diff_epoch}
  \end{table}

\section{Result}
\hspace{-5pt}
We analyse the influential factors of the model's outputs, including prompt styles and different hyperparameters, as well as the generation errors and potential reasons.
\subsection{Hyperparameter Sensitivity Analysis}
The experimental results reveal several important trends in how the model responds to different training strategies and hyperparameters. First, Table~\ref{tab:diff_strategy} demonstrates that each adjustment to decoding parameters or prompt prefixing contributes progressively to narrowing the gap between generated sentences and reference labels. The single largest drop in edit distance of -4.16 comes from the Standard-to-Adjusted step. This phenomenon is consistent with the modified decoding settings in "$model.generate$", especially the increased beam width that allows the model to explore more possible high-quality continuations and improve the chance of selecting closer answers with reference sentences. Other parameters also prevent the outputs from redundant or looping phrases and repetitive bigram tokens. Additionally, the Adjusted-with-Length one yields the best performance across all metrics. The greatest improvement of BLEU (+0.07) also occurs in this step when changing from the Prefixed solution. Building on the strengths of all previous strategies, it further addresses the issue of sequence-length mismatches. Taken together, these parameter adjustments lead to more concise, fluent generation and significantly enhance the model's ability to parse logical forms and convert them into natural language.

Moreover, Table~\ref{tab:diff_lr} shows that the model with a learning rate of $1e-4$ significantly outperforms the case of $1e-5$ for all metrics. In Table~\ref{tab:diff_epoch}, we observe a steady gain when extending from 20 epochs to 25, and the performance peaks at 30 epochs with only marginal progress. Overall, these findings suggest that an “Adjusted with Length” strategy with $1e-4$ as the learning rate provides the best output, while the number of epochs only has a slight effect on final results. Under this configuration on T5-large with 20 epochs, it achieved optimal performance across all average metrics with an edit distance of 20.17, a normalised score of 1.46 and a BLEU of 0.67.

\subsection{Error Analysis}
We manually examine the first five candidate-reference pairs in the Validation set. According to Appendix Figure~\ref{tag:eg_Val}, the model performs perfectly on simple FOL formulas with only one predicate in examples 1, 4 and 5. However, when the logical templates become more complex, especially with multiple quantifiers or nested clauses, it struggle with two main errors. First, the model omits the intended contrastive relationship. In example 2, it does not map the exclusive logic to the corresponding connective “$but$”. Second, it tends to simplify noun phrases by dropping multi-word modifiers. The “$harmful\ things$” are reduced to “$harm$” in example 3. Moreover, example 3 shows the difference in letter cases, where the model replaces “$Acts$” with “$acts$”. Whether this should be regarded as an error is controversial, as “$acts$” is the correct lowercase form unless “$Acts$” represents a special term. Together, the model correctly captures the main logical relationships from the original formulas but fails to translate more precise semantic meaning.

Additionally, Appendix Figure~\ref{tag:eg_Test} shows that nearly all sentences generated by the model are grammatically correct English and read smoothly. It indicates that the model has the generalization ability to handle simple quantifiers and logical connectives and correctly map them in the out-of-domain dataset. However, when it comes to more complex patterns with multiple indicators, many outputs collapse the nested forms into simplified statements, missing the original logical scope. Less than 50\% of examples fully preserve logical structures. Also, some outputs appear to have strange semantic meanings. The cases involving the XOR operator ($\oplus$) are frequently misunderstood as the meaning of "or". Consequently, the initial exclusive operator is reduced to parallel clauses, losing the critical distinction between exclusive and inclusive disjunctions. For this inconsistent processing, a more precise definition may be needed to make the model understand when to substitute the original symbol for English phrases.

\section{Discussion}
The T5 model on logic-to-text tasks achieves strong performance on in-domain distribution and can convert the synthetic logical forms into fluent natural sentences. Despite these strengths, there remain several limitations. 

First, my observation reveals that as quantifier depth gradually increases, sentences generated by the model become unnatural and affect the final translation quality. As a famous idea coming from Chomsky, human language allows for infinite recursion, but people can not produce or understand unlimited recursion due to memory constraints. Similarly, language models should be able to process nesting data up to their maximum layer depth, but struggle with converting extreme nested logical forms in practice.

Second, the semantic coherence of outputs heavily depends on the coverage and alignment of the predicate and entity vocabularies in Module B’s test-set construction. Further research can consider prompt engineering in LLMs to select optimal predicate–entity pairs, thereby producing natural sentences with semantic richness and more real-world knowledge. This limitation also raises the ethical concerns of FOL2NS to some extent. If certain predicates or entities only appear in specific contexts, the model may learn to generate biased descriptions. Future work should involve a double-check of all representations within the framework. Third, this study lacks true labels for the synthetic test set. It would be a wise choice to quantify the model’s outputs by human evaluation or additional reference annotations. It can also replace validation loss with the BLEU score as an early stopping checkpoint to provide better task-oriented performance in the future. 

\section{Conclusion}
For logical-to-text translation, FOL2NS ensures both formally correct and semantically diverse outputs in general. The model in the final stage achieves low edit distance and high BLEU scores on in-domain datasets, and maintains fluency when generating texts from synthetic logical forms, though it shows limitations in fine-grained semantics and extreme nesting. Future work will focus on prompt engineering to dynamically select predicate-variable pairs, better training and audit strategies, and human-annotated evaluation. 

\bibliography{custom}

\clearpage
\onecolumn
\appendix
\section{Related Figures}
\begin{figure*}[htbp]
\centering
\includegraphics[width=0.88\linewidth]{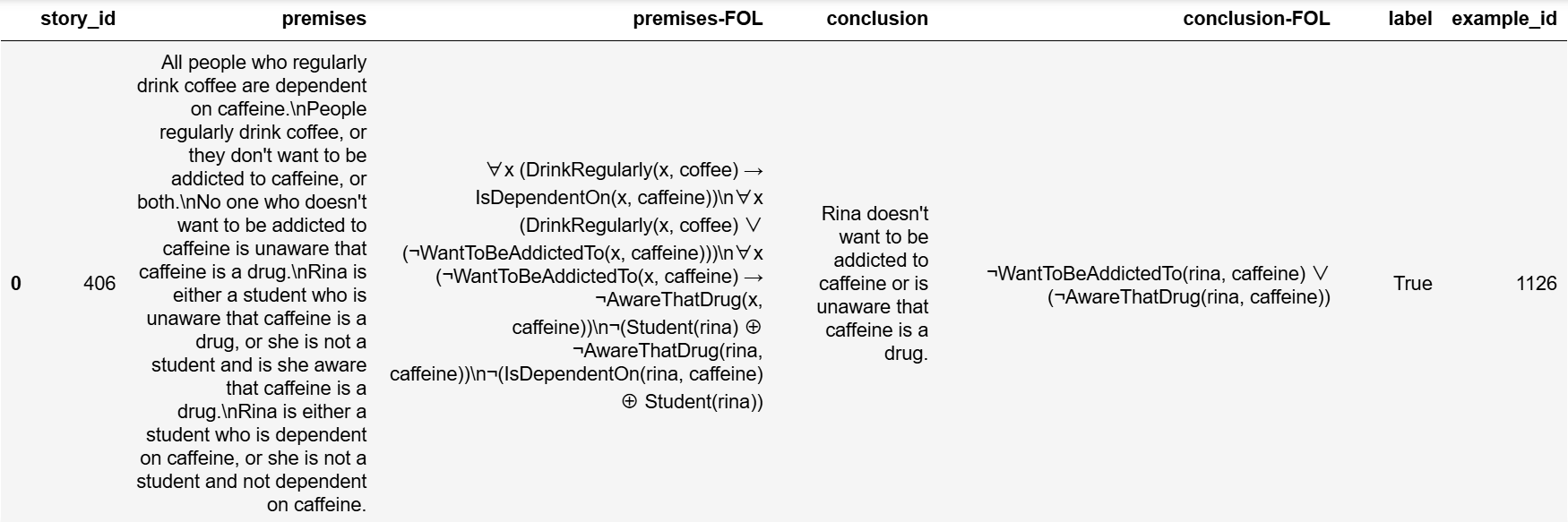} 
\caption {An example of FOLIO.}
\label{tag:FOLIO_eg}
\end{figure*}

\begin{figure*}[htbp]
\centering
\includegraphics[width=0.4\linewidth]{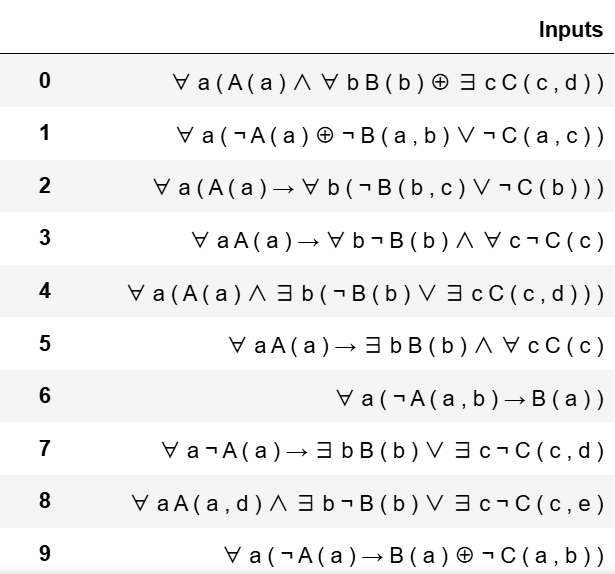} 
\caption {The first 10 examples of defined FOL formulas in Module B.}
\label{tag:eg_FOL}
\end{figure*}

\begin{figure*}[htbp]
\centering
\includegraphics[width=0.88\linewidth]{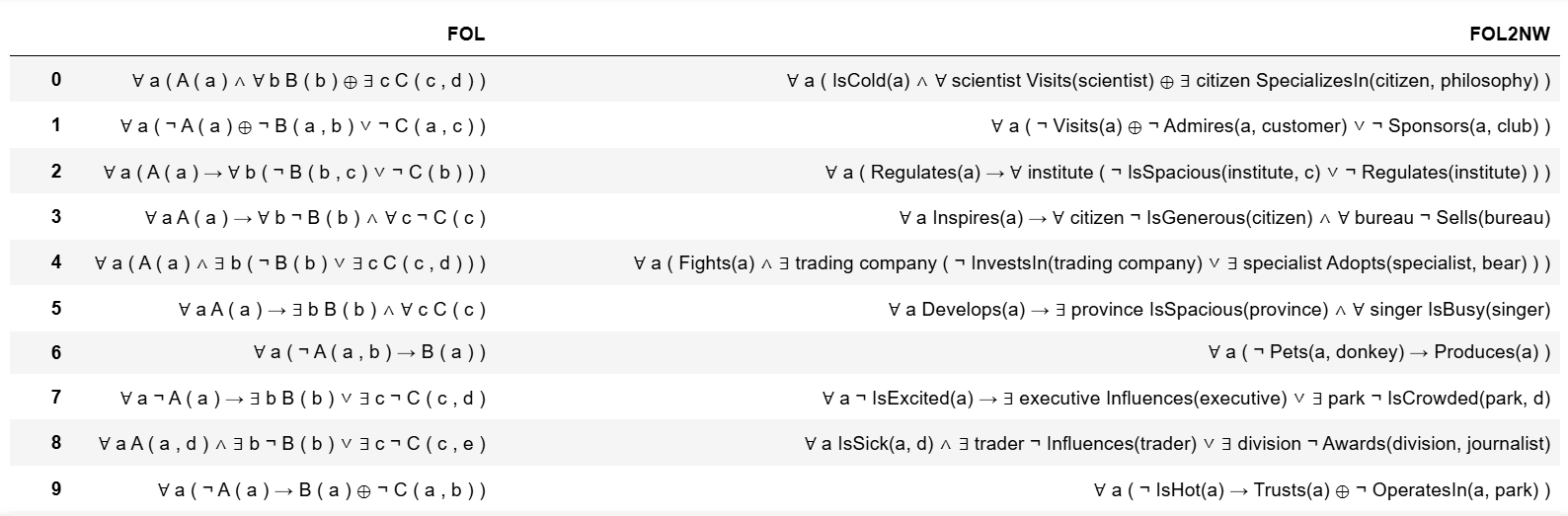} 
\caption {Examples of the preprocessed FOL format with selected predicate-variable pairs (FOL2NW).}
\label{tag:Pv_pairs}
\end{figure*}

\begin{figure*}[htbp]
\includegraphics[width=1\linewidth]{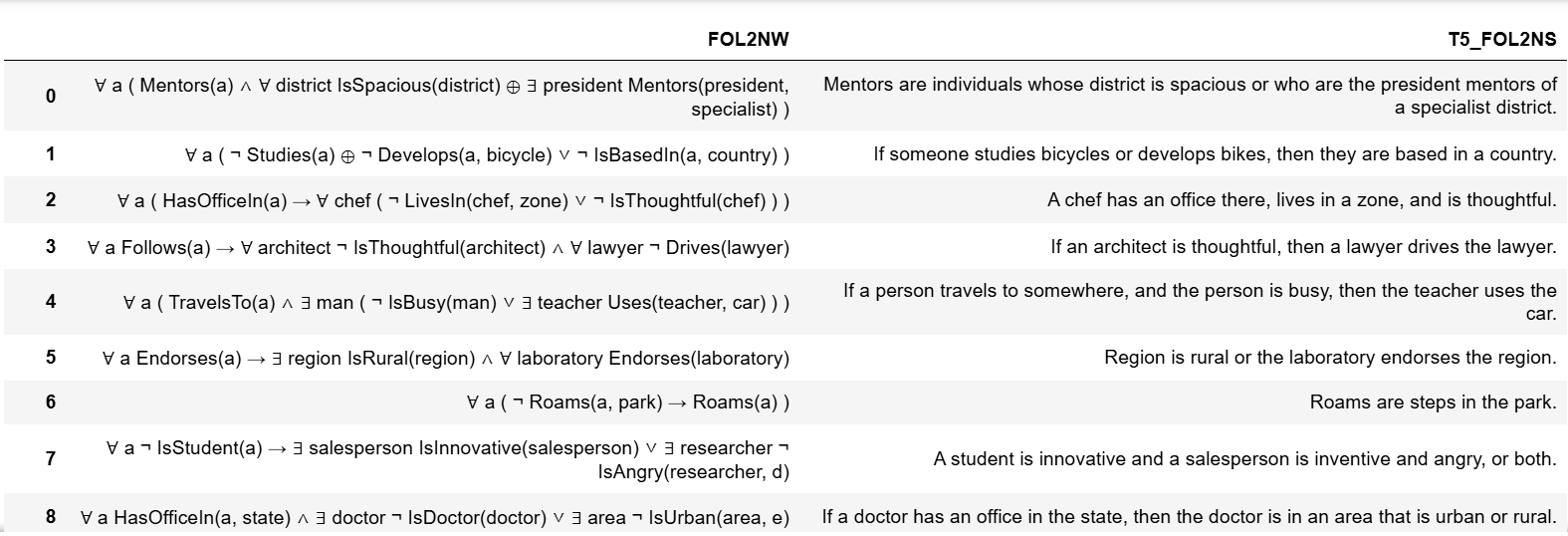} 
\caption {Examples of T5-generated translations (T5\_FOL2NS) from the FOL2NW stage.}
\label{tag:T5_result}
\end{figure*}

\begin{figure*}[htbp]
\includegraphics[width=1\linewidth]{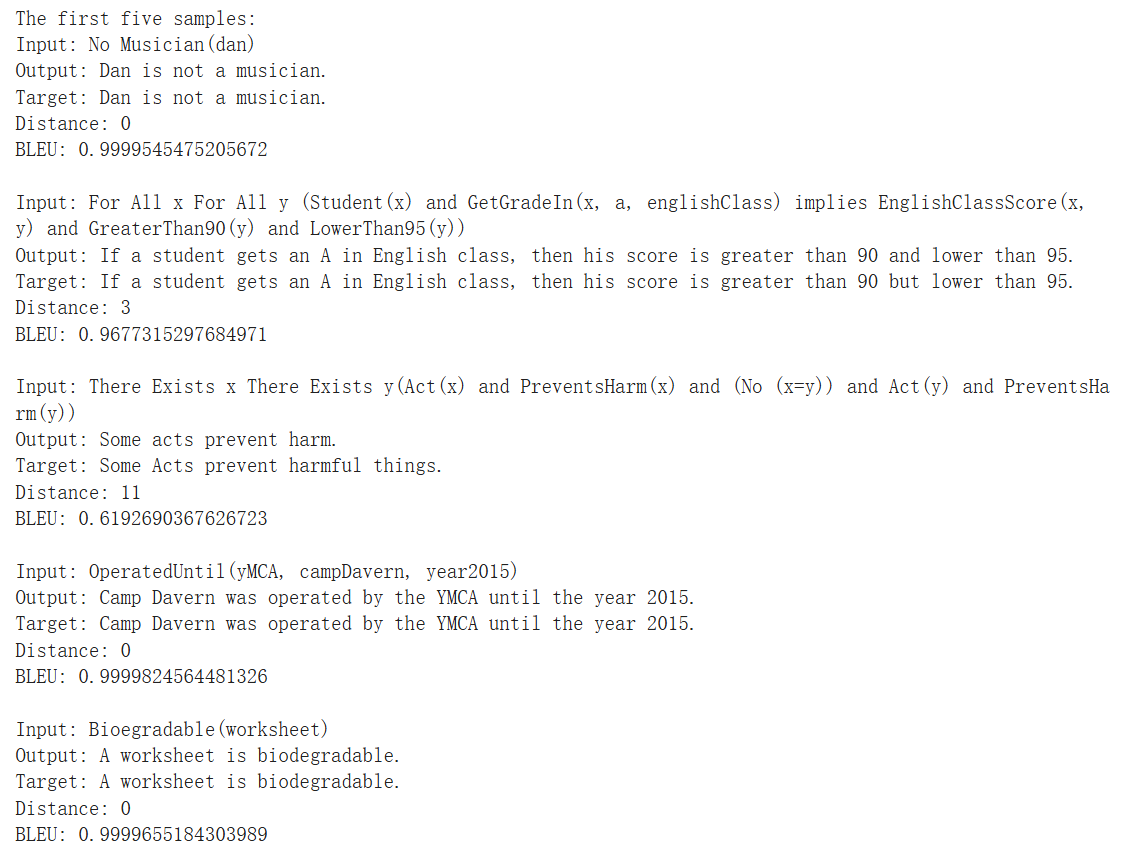} 
\caption {Part of the model results of inputs (logical form), outputs (candidate) and targets (reference) in the Validation set.}
\label{tag:eg_Val}
\end{figure*}

\begin{figure*}[htbp]
\includegraphics[width=1\linewidth]{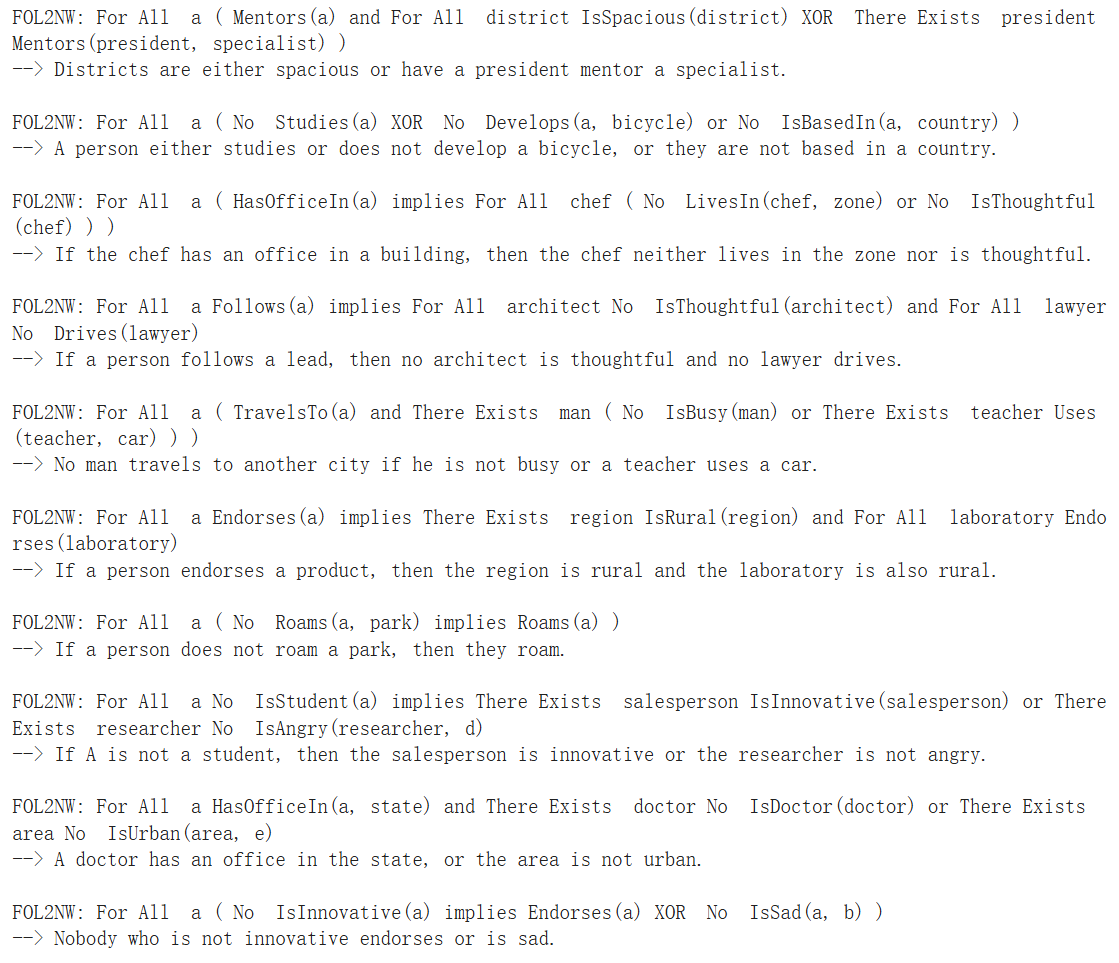} 
\caption {Part of the model results of inputs (logical form) and outputs (candidate) in the Test set.}
\label{tag:eg_Test}
\end{figure*}

Figure~\ref{tag:FOLIO_eg} shows examples of FOLIO, and Figure~\ref{tag:eg_FOL}, Figure~\ref{tag:Pv_pairs} and Figure~\ref{tag:T5_result} present examples from the full generation stages of FOL2NS datasets, including original FOL formulas, FOL-to-FOL2NW mapping and FOL2NW-to-FOL2NS translation by the T5 model. Figure~\ref{tag:eg_Val} and Figure~\ref{tag:eg_Test} are related to the generated outputs of T5 on the Validation set and Test set.
\vspace{30pt}
\vspace{-5pt}
\vspace{-5pt}

\end{document}